\documentclass[journal,12pt]{IEEEtran}
\usepackage{cite}
\usepackage{amsmath,amssymb,amsfonts}
\usepackage{algorithmic}
\usepackage{graphicx}
\usepackage{textcomp}
\usepackage{xcolor}
\usepackage[capitalise]{cleveref}
\usepackage{amsmath,bm}
\usepackage{mathtools}
\usepackage{amsthm}
\usepackage{tabularray}
\usepackage{booktabs, makecell, multirow, tabularx}
\usepackage{url}
\usepackage{threeparttable}
\usepackage[tight,footnotesize]{subfigure}

\usepackage[caption=false,font=footnotesize]{subfig}

\theoremstyle{plain}
\newtheorem{theorem}{Theorem}[section]

\newtheorem{definition}{Definition}[section]

\theoremstyle{remark}

\Crefname{definition}{Definition}{Definition}
\begin{document}

\title{Enhancing Privacy in ControlNet and Stable Diffusion via Split Learning}
\author{Dixi~Yao
	\IEEEcompsocitemizethanks{\IEEEcompsocthanksitem University of Toronto, Toronto, ON M5S 1A1, Canada. E-mail: dixi.yao@mail.utoronto.ca}}
\maketitle

\begin{abstract}
    With the emerging trend of large generative models, ControlNet is introduced to enable users to fine-tune pre-trained models with their own data for various use cases. A natural question arises: how can we train ControlNet models while ensuring users' data privacy across distributed devices? Exploring different distributed training schemes, we find conventional federated learning and split learning unsuitable. Instead, we propose a new distributed learning structure that eliminates the need for the server to send gradients back. Through a comprehensive evaluation of existing threats, we discover that in the context of training ControlNet with split learning, most existing attacks are ineffective, except for two mentioned in previous literature. To counter these threats, we leverage the properties of diffusion models and design a new timestep sampling policy during forward processes. We further propose a privacy-preserving activation function and a method to prevent private text prompts from leaving clients, tailored for image generation with diffusion models. Our experimental results demonstrate that our algorithms and systems greatly enhance the efficiency of distributed training for ControlNet while ensuring users' data privacy without compromising image generation quality.
\end{abstract}

\section{Introduction}

Leading at the forefront in the emerging trend of large generative artificial intelligence, large diffusion models~\cite{diffusionmodel} have become commercial success stories, with models from Stability AI and Midjourney dominating the news. With large diffusion models, any user is able to generate artistically appealing images with short descriptive text prompts. However, short descriptive text prompts do not offer a sufficient level of control over the generated images to satisfy a user's needs in many cases. To support an additional level of control using \emph{conditions}, ControlNet~\cite{controlnet} has recently emerged, allowing users to generate images with a wide variety of user-defined conditions beyond text prompts.

With fine-grained control over generated images using ControlNet, it's intuitive that users would want to fine-tune pre-trained ControlNet models with their own data to meet various use cases. However, since the training dataset may contain users' own artistic creations or faces, privacy concerns arise. Additionally, each user may possess only a small number of images, which may not suffice for fine-tuning a diffusion model unless aggregated, such as in a collection of \emph{50,000} images~\cite{controlnet}. To maintain data privacy, it's essential to fine-tune ControlNet with distributed users, posing the research question: How can we train ControlNet models while preserving users' data privacy, particularly when the data is distributed across multiple client devices?

Federated learning has been heralded in recent years as a distributed training paradigm that preserves user privacy by training directly on client devices and aggregating local training updates using a federated learning server. However, conventional federated learning may not be suitable for fine-tuning large ControlNet and stable diffusion models for three important reasons. \emph{First}, ControlNets and stable diffusion models are large generative models, requiring formidable GPU resources on client devices for local fine-tuning of pre-trained models. \emph{Second}, even if such GPU resources were available on client devices, pre-trained ControlNet and stable diffusion models may not be accessible as open-source due to commercial interests. For example, neither OpenAI nor Midjourney has open-sourced models such as DALL$\cdot$E 2~\cite{ramesh2022hierarchical}. \emph{Finally}, our own experimental results, as presented in this paper, indicate that large ControlNet models fine-tuned with conventional federated averaging~\cite{fedavg} as the aggregation mechanism experienced severely degraded performance compared to centralized training.

Hence, \emph{split learning}\cite{split-learning} becomes the only feasible distributed training paradigm for fine-tuning ControlNets. Clients train the first few layers of the neural network with their local data and transmit \emph{intermediate features} to the server. The server then sequentially sends gradients back to clients after the forward pass and backpropagation. However, recent literature highlights that split learning can be inefficient and vulnerable to adversarial attacks, such as inversion attacks\cite{xiao2020adversarial, patchshuffling, datamix, unsplit}, which have the potential to reconstruct private data.

To preserve data privacy, before designing defense mechanisms for training ControlNet in split learning, we begin to have second thoughts on whether these attacks are practical in real-world use cases of training ControlNet and stable diffusion models. We test these existing attacks in practical settings and under valid assumptions and surprisingly discover that images reconstructed by most existing attacks are not recognizable by humans.

With our detailed analysis of existing attacks and case studies, we find that only inversion attacks using inverse network models are effective for reconstructing conditional images when we train models with split learning. These attacks first train an inverse network on a public dataset and then use it to reconstruct private data~\cite{xiao2020adversarial, patchshuffling}. We empirically demonstrate that the success of the attack depends on the types of private data and should be analyzed on a case-by-case basis. Furthermore, we find that defending against such successful attacks with existing defense mechanisms greatly degrades image generation performance.

Our original contributions are as follows:

\emph{First}, to enhance the efficiency of fine-tuning ControlNets using split learning, we design a new deployment structure. This structure eliminates the need for the server to send data back to the clients, thereby addressing the issue of efficiency bottlenecks.

\emph{Second}, inspired by our empirical observations, we find that the forward process when training diffusion models can be combined with local differential privacy guarantees. Based on this, we emphasize our privacy-preserving timestep scheduling policy, establishing a relationship between the timestep scheduling policy and the privacy budget $\epsilon$. This allows us to adjust the privacy-preserving ability of the system by setting specific scheduling policies. Additionally, we propose a symmetric activation function to process intermediate features, preventing attackers from reconstructing conditional images while still enabling the generation of high-quality images.

\emph{Third}, in addition to the privacy leakage of conditional images, we further explore the leakage of text prompts. To train the stable diffusion model and ControlNet, we need to upload the text prompts, which may contain private information, to the server. We propose a new mechanism to train ControlNets with zero prompts. This trained model can still maintain high performance in image generation, while the server does not know the text prompts.

\emph{Finally}, to evaluate performance in production settings, we implement a system to train ControlNet with federated learning and split learning using \textsc{Plato} and conduct experiments in real-world settings. It is demonstrated that with split learning and our architecture design, clients require less than 3 GB of GPU memory and experience $3 \times$ lower communication overhead. Unlike existing privacy-preserving mechanisms, we verify that our mechanism can protect the privacy of images, conditions, and text prompts without sacrificing image generation performance.

\section{Background and Related Work}
\begin{figure}[tb]
 \centering
 \includegraphics[width=\linewidth]{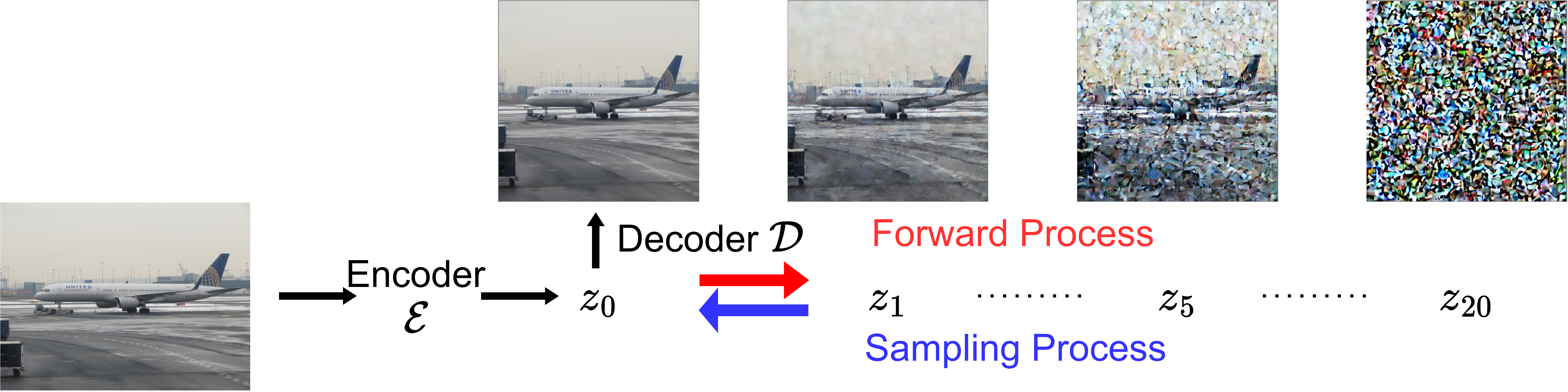}
 \caption{Examples of images output from the decoder corresponding to latent representation at different timesteps during forward and sampling process.}
 \label{fig:flowdiffusion}
\end{figure}
\subsection{Diffusion Model and ControlNet}
Diffusion Models~\cite{diffusionmodel} are probabilistic models used to learn a data distribution by gradually denoising a normally distributed variable to generate high-quality images. The stable diffusion model~\cite{SDV2} converts images into latent representations with an encoder $\mathcal{E}$ and conducts the diffusion process on the latent domain $Z$. After the sampling process, images are generated through a corresponding decoder $\mathcal{D}$.

Existing diffusion models allow users to guide image generation with text prompts. For example, in a stable diffusion model, we utilize a contrastive language-image pretraining (CLIP) model~\cite{clip}, with parameters $\gamma_\theta$, to convert text prompts into features. A cross-attention layer is then employed to combine these features with latent representations.

The image generation process involves a sampling procedure, which is the inverse of the forward process depicted in \cref{fig:flowdiffusion}. In the forward process, we follow a Markov Chain to gradually add Gaussian noise ($\mathcal{N}\sim(0,1)$) to the data, based on a variance schedule $\beta_1,\ldots, \beta_T$, where $t\in[T]$ represents each timestep of noise addition. We denote this Gaussian noise as $\hat{n}$. As an inverse of the diffusion process, during the sampling process, the diffusion model outputs an estimation of noise $n$ at timestep $t$, and we sample the latent $z_{t-1}$ using the equation:

\begin{equation}
 \label{eqa:orgformulation}
 \begin{aligned}
 z_{t-1}=&\sqrt{\alpha_{t-1}}\left(\frac{z_t-\sqrt{1-\alpha_t}n}{\sqrt{\alpha_t}}\right)\\
 &+\sqrt{1-\alpha_{t-1}-\lambda_t^2}n+\lambda_t o(z_t)
 \end{aligned}
\end{equation}

Here, $\alpha_t=1-\beta_t$, $\lambda_t$ is a noise coefficient, and $o(z_t)$ is a small value randomly generated from a standard normal distribution. The sampling process begins with randomized Gaussian noise and gradually samples until we obtain $z_0$, which corresponds to the latent representation of the image we wish to generate.

During the training process, we uniformly sample timesteps from $[T]$ and train the diffusion model (DM) with given text prompts $y$, aiming to minimize the loss:

\begin{equation}
 \min \mathcal{L}_{\rm DM}=\mathbb{E}_{\mathcal{E}(x),y,n\sim\mathcal{N}(0,1),t}\left[\lVert \hat{n}-n_\theta(z_t,t,\gamma_\theta(y))\rVert^2_2\right]
\end{equation}

In each training step, we need to follow \cref{eqa:orgformulation} to generate a random noise $\hat{n}$ as a label, which serves as the ground truth. The diffusion model's objective is to learn the parameters $\theta$, which enable it to infer the noise $n$. This inferred noise is the output of the diffusion model, used for denoising the image.

\begin{figure}[tb]
 \begin{minipage}{0.48\linewidth}
 \includegraphics[width=0.48\linewidth]{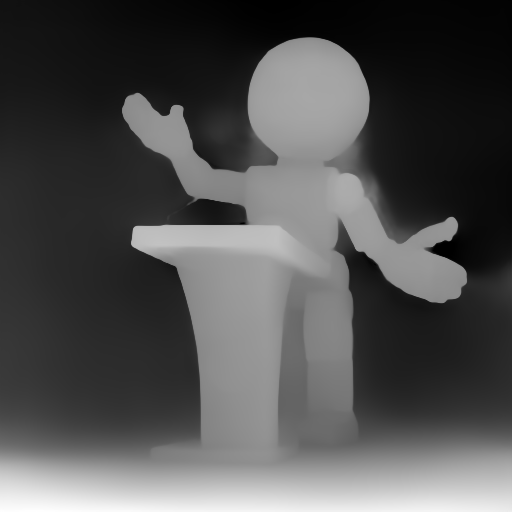}\hfill
 \includegraphics[width=0.48\linewidth]{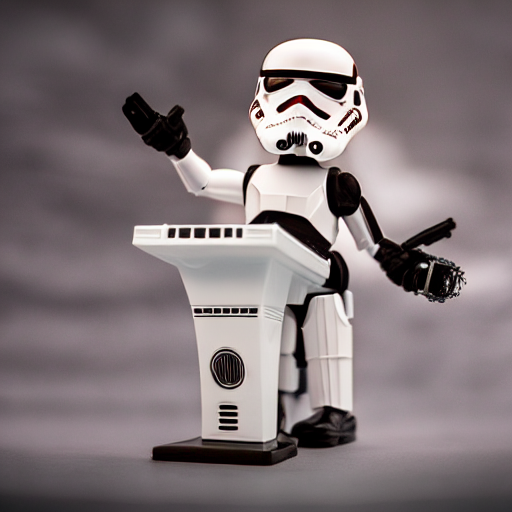}
 
 \caption{The image on the right is generated from the ControlNet with a condition image on the left. The condition image is an image of depth maps. The text prompt is: \emph{Stormtrooper's lecture}.}
 \label{fig:controlnetorg}
 \end{minipage}
 \hfill
 \begin{minipage}{0.48\linewidth}
 \includegraphics[width=0.48\linewidth]{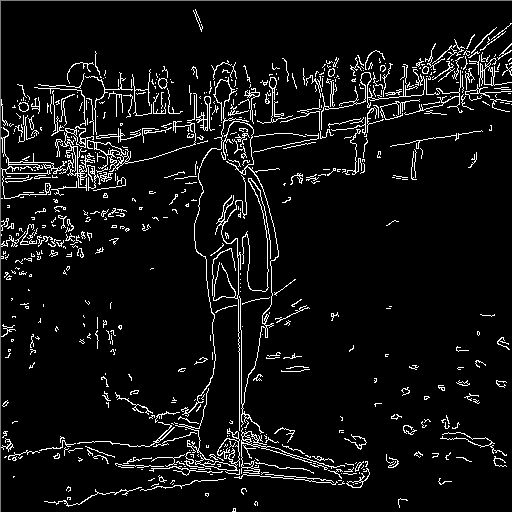}\hfill
 \includegraphics[width=0.48\linewidth]{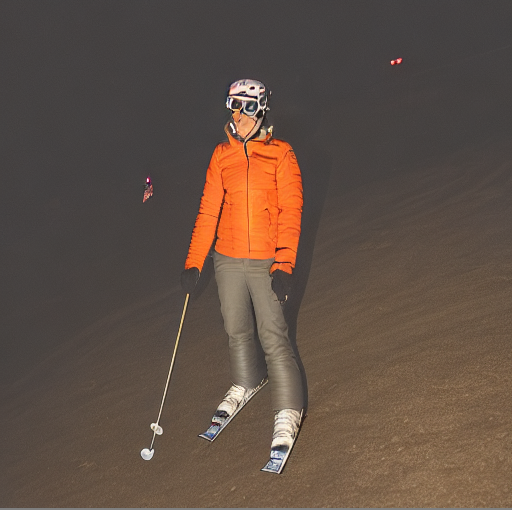}
 
 \caption{The image on the right is generated from the ControlNet trained by FedAvg using the condition image on the left. The text prompt is: \emph{A skier poses for a shot on the night time slopes}.}
 \label{fig:flfigure}
 \end{minipage}
\end{figure}

To enable users to control the generated images with more detailed conditions such as scribbles~\cite{controlnet}, canny lines~\cite{canny1986computational}, depth maps~\cite{ranftl2020towards}, HED lines~\cite{xie2015holistically}, and segmentation maps~\cite{zhou2017scene}, in addition to the given text prompts, a conditional diffusion model is proposed. An example is ControlNet~\cite{controlnet}, as shown in \cref{fig:controlnetorg}. We can generate a stormtrooper with the same skeletons as in the left image of the depth maps.

ControlNet copies the encoders from the backbone diffusion models and replaces the decoders of the backbone diffusion models with convolution layers initialized with zeros (referred to as zero convolution). A control network comprises copied encoders, zero convolutions, and a condition encoder for converting conditions into latent representations. ControlNet consists of this control network and the original stable diffusion model. For the detailed structure of ControlNet, we will explain it later in \cref{sec:sl}.

Apart from stable diffusion models, ControlNet can also leverage other backbones such as LCM~\cite{luo2023latent} and ControlLoRA~\cite{wu2023controllorav2}. Concurrent works, T2I-Adapter~\cite{t2i} and Composer~\cite{huang2023composer}, feature much smaller and much larger control networks, respectively. The control network can have other structures, such as those seen in T2IAdapter~\cite{t2i} and Composer~\cite{huang2023composer}. Our method can also be applied to these works. FreeDoM~\cite{yu2023freedom} is a training-free conditional diffusion model. However, generating images with fine-grained conditions, such as using canny edge maps, can be challenging, resulting in poor guidance. Training-required methods still remain the optimal solution for conditional diffusion models.

\subsection{Decentralized ControlNet Training}
With the assistance of ControlNet, users can fine-tune well-trained stable diffusion (SD) models without disrupting the original SD models. However, the conditions and training images involved may contain privacy-sensitive information. One straightforward solution is to train ControlNet entirely on a single device. For inference with a batch size of 1, we need 7.50 GB of GPU memory. However, to train ControlNet, a minimum of 23.82 GB of GPU memory is needed (with a minimal batch size of 2). Such high GPU memory requirements are unfeasible for most client devices, let alone mobile devices. Even if clients possess powerful computing resources and acceptable training times, training ControlNet on the client side remains impossible if the server is unwilling to share the well-trained diffusion model.

Even if a client has enough GPU memory to fine-tune a diffusion model locally, another issue arises when it needs to collect training samples from different users, as it may lack sufficient samples in its local data. To allow users to fine-tune ControlNets without their private data leaving their devices, a common solution is to leverage privacy-preserving decentralized frameworks. One such decentralized training paradigm is federated learning \cite{fedavg,yang2023privatefl}. We follow the standard federated averaging scheme to train the ControlNet with 50 clients, each having 1000 training samples. We train for a total of 100 rounds and aggregate weights after every 250 local iterations. We evaluate the performance on the MS-COCO \cite{coco} validation set. As shown in \cref{fig:flfigure}, even under the assumption that clients have powerful computing units and the weights of the diffusion model are available, the ControlNet trained by FedAvg \cite{fedavg} fails to learn the conditions. The generated image does not match the condition at all; for example, the posture of the person in the generated image differs from that in the left condition images. Since federated learning without any privacy-preserving mechanism cannot work, we do not need to further evaluate methods with privacy-preserving mechanisms \cite{yang2023privatefl}.

There are also data encryption approaches in decentralized systems, such as trusted execution environments, multi-party computation, and homomorphic encryption. However, the overhead is not at the same scale as computing over plaintext data. For example, during inference, the forward time on diffusion models with homomorphic encryption~\cite{chen2024privacypreserving} is 79.19 days, compared to 35 seconds with plaintext using NVIDIA A100. Moreover, to the best of our knowledge, there is no encryption method that can be directly applied to the training process of diffusion models.

Considering the challenges involved in training conditional diffusion models either on clients or servers, a suitable solution is to train such models through split learning, involving multiple clients and the server. Unlike federated learning, which pushes the entire model to the edge, split learning employs a neural network spanning both the cloud and the edge. An edge device trains the network up to the partition layer and sends the intermediate features to the server. Upon receiving these features, the server takes over training the remaining layers and completes forward propagation. During backward propagation, the server conducts back-propagation up to the partition point and sends the gradients of the partition layers back to the client. The client then updates the local parameters through back-propagation using the received gradients. A major drawback of split learning is its sequential training manner, resulting in significant resource underutilization and high transmission overhead, leading to longer training times. In each training step, the server and clients exchange features and gradients, with one party waiting while the other computes or transmits data.

\subsection{Privacy Leakage in Split Learning}
Potential threats arise from split learning, as it carries the risk of privacy leakage through data transmission between clients and the server. Literature highlights that an honest-but-curious server could reconstruct private data using the intermediate features sent from clients to the server. Zhang et al.~\cite{zhang2020secret} successfully reconstructed private data in a white-box setting. UnSplit\cite{unsplit} further refined this method to conduct a similar attack in a black-box setting. He et al.~\cite{he2019model} trained an inverse network using a public dataset, taking intermediate results as inputs to output private data for reconstruction. Conversely, Li et al.~\cite{li2021label} demonstrated that private labels on clients also face the risk of leakage. Pasquini et al.~\cite{pasquini2021unleashing} proposed an attack to reconstruct private data by manipulating gradients sent back to clients, under the assumption of a dishonest server. Duan et al.~\cite{duan2023diffusion} introduced a membership inference attack (MIA) tailored specifically for diffusion models, although they acknowledged its limited applicability in real-world scenarios. Direct MIA is excluded from the scope of our paper. Additionally, Carlini et al.~\cite{nicolas2023extracting} utilized leaked text prompts to generate numerous images, subsequently employing MIA to identify which images exist in private datasets.

\subsection{Privacy Protection in Split Learning}
In response to potential privacy leakage in split learning, researchers have made efforts to defend against such attacks. Local differential privacy techniques, such as additive noise and randomized response~\cite{ldp}, are employed to prevent reconstruction. Additionally, Gaussian noise is utilized to directly add noise to the raw data~\cite{datamix}. Subsequently, many works have adopted methods involving additive noise \cite{titcombe2021practical,vepakomma2019reducing,shredder}. DataMix~\cite{datamix} and CutMix~\cite{oh2022differentially} leverage the concept of mixing a batch of samples. DataMix is designed for convolutional neural networks, while CutMix is tailored for vision image transformers. PatchShuffling~\cite{patchshuffling,xu2023shuffled,xu2023shuffled2} is a method specifically designed for transformer-structured models, where patches are shuffled among a batch of samples. However, all these methods must provide sufficient privacy guarantees at the cost of significant performance decreases.

Xiao et al.~\cite{xiao2020adversarial} utilized adversarial learning to enable clients to generate intermediate results that the server cannot use to reconstruct images. Shredder\cite{shredder} introduced noise based on mutual information, while DISCO \cite{disco} employed a channel obfuscation method to process features before transmitting them to the server. However, all three of these methods are only applied during the inference stage and require training a network to process features.

\section{Preliminaries}
\subsection{Local Differential Privacy}
Differential privacy was originally conceived to ensure that an adversary's ability to compromise the privacy of any set of users remains unchanged by an individual's decision to opt into or out of the dataset~\cite{shokri2015privacy}. This characteristic ensures that an adversary cannot glean additional information about any specific individual, thereby extending its capability to prevent inversion attacks such as reconstructing private data.

We assume we have two sets $X$ and $X'$, with only one sample difference between them. We say a privacy-preserving mechanism $\mathcal{M}$ is locally differentially private (LDP) if we cannot differentiate between $X$ and $X'$ based on the outputs of these two sets by $\mathcal{M}$. We provide a formal definition here.

\begin{definition}
 ($(\epsilon,\Delta)-LDP$) A mechanism $\mathcal{M}$ is $(\epsilon,\Delta)-$LDP if for two adjacent input sets $X$ and $X'$, and a set $\mathcal{O}$ of all possible outputs,
 \begin{equation}
 {\rm Pr}[\mathcal{M}(X)\in \mathcal{O}]\leq e^{\epsilon}\cdot{\rm Pr}[\mathcal{M}(X')\in \mathcal{O}]+\Delta
 \end{equation}
\end{definition}

We call this $\epsilon$ privacy budget. In common cases, a larger privacy budget implies easier differentiation between the two probabilities, indicating weaker privacy-preserving ability, and vice versa.

We can apply a differential privacy mechanism to either model weights or features. DPDM~\cite{dockhorn2023differentially} and DPGM~\cite{jiang2024functional} have implemented differential privacy mechanisms over the model weights of a diffusion model. However, since our focus is on split learning, we need to concentrate on mechanisms applied to the inputs or features. To achieve privacy protection against reconstructing private images, we can add noise to the original inputs or intermediate features~\cite{dwork2008differential}, making it $(\epsilon,\Delta)$-LDP. The noise added can be Gaussian noise \cite{ldp,gdp}.

\begin{definition}
 \label{def:eLDP}
 (($\epsilon,\Delta)-LDP$ noise adding.) A mechanism of adding noise over samples is $\epsilon-$LDP if the Gaussian noise follows the normal distribution
 \begin{equation}
 \mathcal{N}\sim\left(0,2\ln\frac{1.25}{\Delta}\alpha^2\cdot\frac{1}{\epsilon^2}\right)
 \end{equation}
 In literature, $\alpha$ is called sensitivity. It is the biggest $L_2$ distance between all possible inputs or intermediate features we are going to add noise.
\end{definition}

We will later use this formulation of adding noise in the diffusion model to propose an $(\epsilon,\Delta)$-LDP mechanism over features. Randomized response~\cite{ldprr} is another typical mechanism to achieve local differential privacy. It encodes each real value in a feature into bits and randomly flips each bit. A drawback of existing LDP mechanisms, whether applied to model weights or features, is that they aim for privacy preservation at the expense of the quality of generated images.

\subsection{Training ControlNet with Split Learning}
\label{sec:sl}
\begin{figure}[tb]
 \centerline{\includegraphics[width=\linewidth]{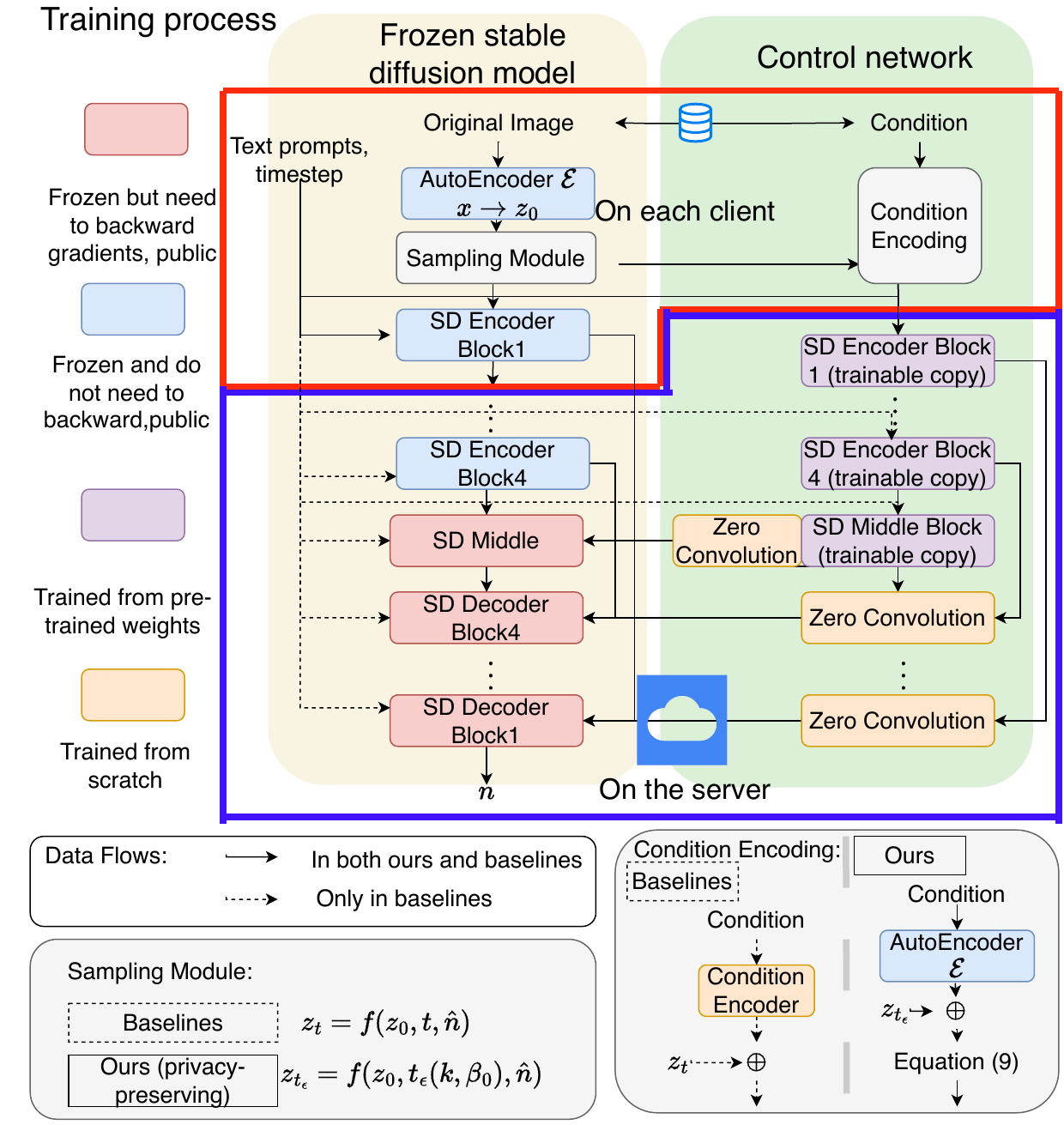}}
 \caption{The deployment of ControlNet across the clients and the server under split learning framework and our proposed privacy-preserving framework during training. Function $f$ is $z_t=f(z_0,t,\hat{n})=\sqrt{\alpha_t}z_0+\sqrt{1-\alpha_t}\hat{n}$.}
 \label{fig:cutcontrol}
\end{figure}

We first introduce the deployment when we want to fine-tune a diffusion model with ControlNet using split learning. Though this is engineering work to combine two existing frameworks, we would still like to provide the essential rationale behind how we decide the cut layers and model deployments over clients and the server. Different from the partition of dense models such as ResNet \cite{resnet}, which has the structure of blocks being sequentially placed, the conditional diffusion model contains two parts: a diffusion model with frozen weights and a trainable control network. The structure of the ControlNet is shown in \cref{fig:cutcontrol}. The control network will first process the condition images with a particular condition encoder (during condition encoding) trained from scratch and then mix it with the noisy latent representation as the input of the following blocks. 

The diffusion model will first use a pre-trained encoder to convert the original image into the latent representation. The diffusion model has encoder blocks and decoder blocks. As they are all frozen, no parameters need to be updated. But the decoder blocks need to calculate gradients of parameters so as to update the control network parameters during backpropagation. Each diffusion model decoder will take the output of the corresponding encoder and the output of the corresponding block in the control network as the inputs. These inputs are fed into the decoder with a jump connection using a similar structure as the UNet \cite{unet}.

For each encoder and decoder block in the original stable diffusion model and each encoder block in the control network, we will also put the text prompts and timestep as the inputs to realize the text-to-image generation. The condition encoder and the autoencoder $\mathcal{E}$ only take the conditions or original images as the inputs. During the forward process, the clients need to send text prompts as well as timesteps to the server.

Considering hiding the complete model weights of the well-trained diffusion models from the clients and achieving the best tradeoff between privacy and efficiency through choosing different partition points, we cut right after the first diffusion model encoder and the trainable condition encoder. If we cut deeper, the server still needs the output of the previous encoder blocks as the inputs of the decoders. This will not provide a better privacy guarantee but will increase the transmission and computation overhead on the clients. 

We send the output of the model and gradients back to the clients In such a way, the images are generated on the clients. The output is the $n$ going to be used in \cref{eqa:orgformulation} for denoising. It is an inferred random noise following the standard normal distribution. The server will not be able to generate images with only knowing $n$. Regarding the diffusion model, if we place the partition point before the first encoder block, the server can subtract the estimated noise $n$ from received $z_t$ in \cref{eqa:orgformulation} to recover the $z_0$ and retrieve the private images.

\section{Speed Up by Not Sending Gradients Back}
\label{sec:ours}
\begin{table}[tb]
 \caption{
 Comparison of memory usage, training time, and transmission overhead for different split learning structures training ControlNet. ${\rm M_c} $ and ${\rm M_s} $ denote GPU memory usage (in GB) on the client and server, respectively. ${\rm T_c} $ and ${\rm T_s} $ indicate training time (in hours) on the client and server. ${\rm T}$ represents transmission overhead (in GB).}
 \label{tab:efficiency}
 \begin{center}
 \begin{tabular}{cccccc}
 \hline
 Structure&${\rm M_c} $& ${\rm M_s} $& ${\rm T_c}$& ${\rm T_s}$& ${\rm T}$\\
 \hline
 Split learning & 2.78&22.04&22.46&14.10&559.17\\
 Ours &2.75&22.04&0.446&14.10&186.56\\
 \hline
 \end{tabular}
 \end{center}
\end{table}

To do split learning in practical use cases, we propose a new deployment structure to address the efficiency bottleneck. This design ensures that the server does not need to send back gradients, thereby removing the sequential dependency between clients and the server during training. Instead of training a condition encoder for each different condition, we propose to replace it with the pre-trained encoder used in the stable diffusion model. This way, clients only need to perform inference, allowing them to continuously forward without waiting for gradients from the server. This approach addresses the bottleneck caused by the sequential training manner.

As the clients share the same pre-trained model and the server model is shared between all clients, we do not need to aggregate client models. This makes the trained ControlNet have the same performance as centralized training. Besides that, since the condition encoder and pre-trained encoder both only need images as inputs, the replacement will not cause the outputs to have distribution drift. Hence, image generation performance will not be affected.

We compare the memory usage, training efficiency, and transmission overhead of these two structures in \cref{tab:efficiency}. The diffusion model is stable diffusion V-1.5, ControlNet is of version 1.1, and the autoencoder is ViT-Large-Patch14 CLIP model~\cite{clip}. The input resolution is $512\times512$. The NVIDIA A100 serves as the server's device, and the NVIDIA A4500 is used for clients' devices. The training batch size is 4, and the model is trained for $2.5\times10^3$ iterations. The number of clients is set the same as in federated learning, which is 50, with each client having 1000 training samples.

Without sending back the gradients, our new structure can save much transmission overhead. Additionally, by eliminating the forward-backward lock between the client and the server, the clients, server, and intermediate data transmission can operate in a parallel pipeline. Clients no longer need to wait for other clients or the server, reducing the time required for each client. Our whole training time is $\max (\{\rm T_c, T_s, T/r\})$ while the original split learning structure needs ${\rm T_c+T_s+T/r}$ for training, where $r$ is the data transmission rate. We can increase the number of clients if we want, but since $T_s$ is much larger than $T_c$, the whole training time is the same.
\section{Re-evaluating Potential Attacks}
\label{sec:privacy}
\subsection{Potential Threats in Split Learning}
\begin{figure}[tb]
 \centerline{\includegraphics[width=\linewidth]{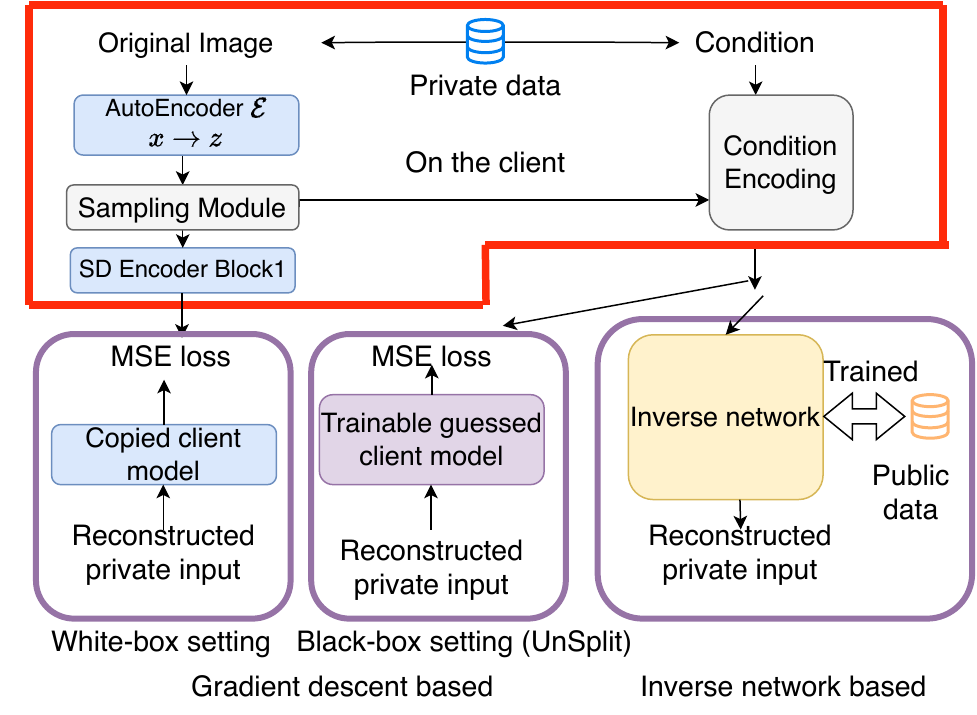}}
 \caption{The illustration of different inversion attacks.}
 \label{fig:attacks}
\end{figure}

\subsubsection{Threat modeling}
We begin by defining the threat model in practical scenarios. We assume the server to be honest but curious. In our designed split-learning structure, although the server does not need to send gradients back to the clients, it will still accurately complete the remaining training in each split learning iteration and send $n$ to the clients. However, simultaneously, the server will attempt to reconstruct private data using the received intermediate features. The server can conduct the reconstruction process in the background, ensuring that clients remain unaware of the attacks.

In our evaluation, we do not consider clients to be malicious. In split learning, a client receives no data if using our proposed framework. Therefore, malicious or colluding clients cannot obtain data related to constructing private images from other clients. However, malicious or colluding clients may send maliciously constructed data to the server to launch other attacks, such as harming model utility. Such cases are detectable as the model cannot generate the correct results. As our focus is on adversaries attempting to reconstruct private images, we do not consider that type of threat.

\subsubsection{Attacking methods}
Several threats have been specifically proposed in split learning, ranging from the leakage of inputs to labels. The most threatening attack is the \emph{inversion attack}, which attempts to reconstruct original private data based on the received intermediate feature. We summarize typical inversion attack methods proposed in previous literature in \cref{fig:attacks}. There are two typical methods to do such an attack.

The first method is based on gradient descent. In this type of attack, the adversary first constructs a randomized input or an input with prior knowledge about the private data. This input is then forwarded through a saved client model on the server, and the reconstruction loss (usually MSE loss) between the output from the randomized input and the received intermediate features is minimized. After several iterations of gradient descent, the randomized input will be optimized to resemble the private data, which we consider to be the reconstruction of private data. The attacker can launch these attacks under a white-box setting~\cite{zhang2020secret} if it knows the parameters of client models; otherwise, it operates under the black-box setting.

In the black-box setting, the first method is a query-based attack \cite{he2019model} where the server sends specific designed inputs to the clients and observes the corresponding intermediate feature output. These queries are not included in all necessary transmitted data discussed in this paper. The second attack method, UnSplit \cite{unsplit}, does not require such queries. In the UnSplit attack methodology, a client model replica, denoted as $M$, is initialized on the server along with a training sample represented as $x$. The parameters of the guessed client model are designated as $\theta$. Following the completion of the UnSplit attack, the converged training samples are utilized as the desired reconstruction of private data. During each iteration of split learning, upon receiving intermediate features denoted as $\hat{h}$, the server feeds the training sample into the guessed client model to obtain the output. Subsequently, the server undergoes multiple inner iterations to update $x$ using $\nabla_x L_{\rm MSE}(M_\theta(x), \hat{h})$, followed by several inner iterations to update $\theta$ using $\nabla_\theta L_{\rm MSE}(M_\theta(x),\hat{h})$. These steps are iteratively performed until convergence is achieved.

The second type of inversion attack is based on training an inverse network \cite{patchshuffling,datamix,he2019model}. In this approach, the attacker first trains an inverse network on a public dataset, which is assumed to have a similar distribution as the private dataset. The inverse network takes the intermediate features as inputs and outputs the reconstructed private data. During the training of inverse networks, if it is under a white-box setting, the attacker will directly use known client model weights to train an inverse network. Otherwise, the attacker will first train an estimated client model using known server model weights on the same public dataset and then use this estimated client model to train an inverse network.

Apart from the leakage caused by the most threatening inversion attack, there are other concerns regarding data privacy leakage. Label leakage~\cite{li2021label} assumes that labels contain private information. The server can infer the private label by observing the distribution of gradients that will be sent back to the clients. However, such an attack is only applicable for binary classification tasks in split learning. Inference attacks~\cite{pasquini2021unleashing} steal private data by sending attacker-designed gradients to fool the client models into sending features that can be used by the attacker to reconstruct the private data.

Another potential leakage is the leakage of text prompts. As depicted in \cref{fig:cutcontrol}, during the training of the ControlNet, the server requires inputting the text prompts into the encoders and decoders of both the original stable diffusion model and the control network. Consequently, clients are required to upload their private text prompts to the server. Some may argue that prompts are short descriptive texts, containing limited useful private information. However, the server could utilize the known prompts to extract a training dataset \cite{nicolas2023extracting}. Therefore, clients must keep the prompts confidential from the server.

\subsection{Re-evaluating the Validity of Assumptions}
\subsubsection{The client model weights can be kept secretly} In a white-box setting \cite{zhang2020secret}, the client model weights are known. However, in real-world scenarios, the client does not need to disclose the model weights to the server for split learning to function. Even if an adversary manages to steal the client model weights, clients can simply re-initialize the model with different parameters. During the training process, if the client model is trainable, its weights will change in each iteration, making such an assumption invalid. The only potential vulnerability arises if the client model is a pre-trained model. Since pre-trained weights are typically publicly available on the Internet, such an attack could pose a threat.

\subsubsection{The client can do split learning without providing prior knowledge about private data to the server} In real-world split learning scenarios, the server only requires the client model for training, operating without any knowledge of the private data. In a black-box setting, it is assumed that the adversary possesses prior knowledge about the private data, enabling it to train an inverse network on public data. For instance, Yao et al.~\cite{patchshuffling} employed CelebA \cite{celeba} as the public dataset and LFWA \cite{lfwa} as the private dataset, both containing human faces. However, in practical contexts, the availability of a public dataset exhibiting such a correlation with private datasets remains uncertain.

\subsubsection{The client can reject the query request} In a specific inversion attack, an adversary must query the client model with samples supplied by the server~\cite{he2019model}. However, in the standard split learning setup, clients do not need to respond to any queries from the server; the split learning still works. Therefore, to counter such an attack, clients can simply reject all queries originating from the server. One may argue that the server could construct these queries in a manner resembling gradients, making them indistinguishable to clients. However, with our structure that eliminates the need for gradient back-sending, such concerns are mitigated.

In inference attacks~\cite{pasquini2021unleashing}, if the client model is trained with gradients designed by the attacker, the resulting model will inevitably experience a performance decline. Users can easily detect this degradation in performance and cease using the compromised server. Furthermore, our designed structure offers a straightforward defense against such attacks as we do not need to train client models.

\subsection{Re-evaluating the Effectiveness of Attacks}
In summary, practical applications of split learning face four threats. The first is a potential attack using gradient descents in a white-box scenario, particularly if clients utilize pre-trained weights. The second threat is an UnSplit attack, while the third involves training inverse networks to infer private data without prior knowledge of the data. The fourth threat is the leakage of text prompts.

\begin{figure}[tb]
 \includegraphics[width=0.98\linewidth]{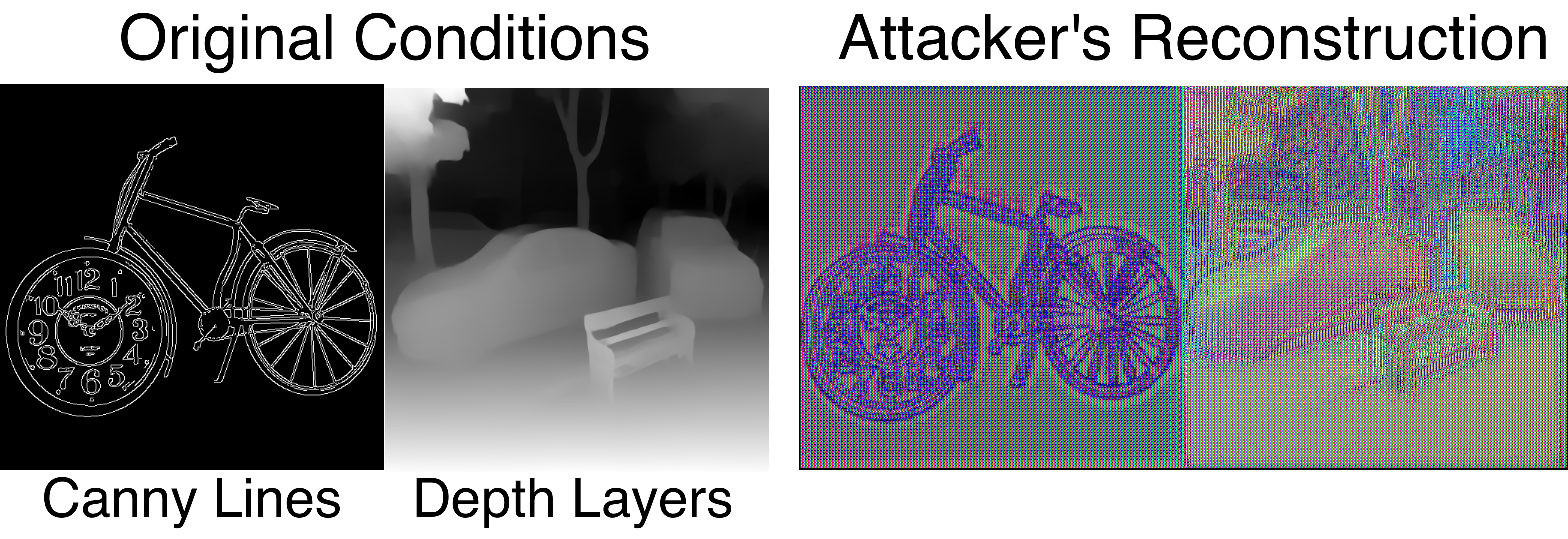}
 \caption{\textbf{Privacy preserving:} Higher distortion means better privacy preservation. Randomly selected and non-cherry-picked examples of reconstructed images by UnSplit attack.}
 \label{fig:unsplit}
\end{figure}

\begin{figure}[tb]
 \includegraphics[width=0.98\linewidth]{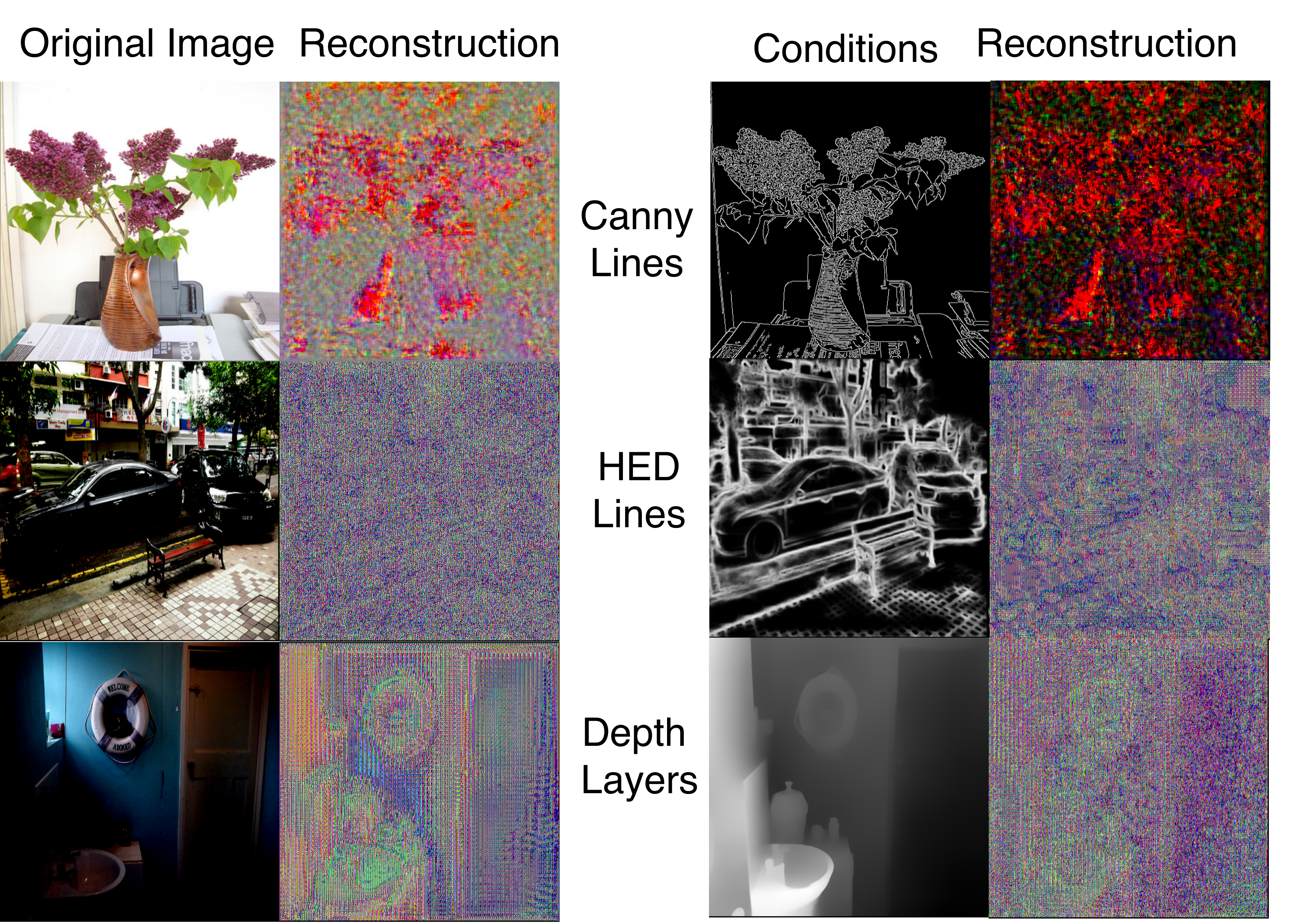}
 \caption{\textbf{Privacy preserving:} Higher distortion means better privacy preservation. Randomly selected and non-cherry-picked examples of reconstructed images by attacks optimizing MSE loss under white-box setting.}
 \label{fig:whitebox}
\end{figure}

\subsubsection{Metrics for privacy-preserving effectiveness}

An honest-but-curious server aims to reconstruct private data based on intermediate results. We evaluate the similarity between reconstructed images and private images using peak signal-to-noise ratio (PSNR) and the structural similarity index measure (SSIM)~\cite{hore2010image}. Private images encompass users' natural and conditional images. Both SSIM and PSNR utilize image pixel values ranging from 0 to 255. PSNR assesses image reconstruction quality, while SSIM gauges image similarity. Lower SSIM and PSNR values signify decreased image similarity, indicating improved privacy preservation.

\subsubsection{Attack by gradient descents} 

In the original structure, since the server lacks knowledge of the condition encoder weights, we resort to the UnSplit attack method, following the procedure outlined in UnSplit~\cite{unsplit}. This attack involves updating the inputs based on the mean squared error (MSE) loss between intermediate results and outputs generated by randomly initialized inputs, iterated over 100 loops. Subsequently, these inputs are used to update the weights of the guessed client model, which is also initialized randomly on the server, for another 100 loops. This process is repeated for a total of 100 outer loops. We optimize the randomized model weights and inputs using the Adam optimizer with a learning rate of 0.001. The loss function utilized is $\mathcal{L}=\mathcal{L}_{MSE}+\mathcal{L}_{L_2}$. For training the ControlNet, the dataset used is MS-COCO \cite{coco}. The successful reconstruction of images using the UnSplit method is illustrated in \cref{fig:unsplit}.

In the gradient back-sending free structure, the server possesses knowledge of the weights of the pre-trained condition encoder. Consequently, the server can launch attacks in a white-box setting. For each attack, we conduct 1000 iterations using the Adam optimizer with a learning rate of 0.001 and MSE as the loss function. Regarding the reconstruction of the original image with the output of the SD encoder block 1, the PSNR is 3.39, and the SSIM is 0.12. For reconstructing the condition image, the PSNR is 5.95, and the SSIM is only 0.002. As depicted in \cref{fig:whitebox}, the reconstructed images are far from recognizable. This ineffectiveness is attributed to the pre-trained autoencoder's complex model structure, which incorporates dropout layers and batch normalization layers. Between the two runs, even with identical inputs, variations in outputs occur due to dropout layers. In dropout layers, the operation of zeroing elements also nullifies the gradient, making methods relying on gradient descent ineffective. Given the ineffectiveness of the white-box setting, we do not need to test the black-box setting Unsplit attack.
\subsubsection{Attack using inverse networks}

\begin{table}[tb]
 \caption{The inverse networks for inversion attacks have two types: Type 1 reconstructs the original image; Type 2 reconstructs the condition image. Type 1\&2 denotes layers used in both structures. Padding size is 1 and kernel size is 3.}
 \label{tab:inverse}
 \begin{center}
 \begin{tabular}{cccccc}
 \hline
 Input&Operator&Stride&\#Out&Structure&Activation\\
 \hline
 $64^2\times320$&Conv2d&1&320&Type 1&SiLU\\
 $64^2\times320$&Conv2d,&1&256&Type 1&SiLU\\
 $64^2\times256$&Upsample&2&96&Type 1&SiLU\\
 \hline
 $64^2\times4$&Upsample&2&96&Type 2&SiLU\\
 \hline
 $128^2\times96$&Conv2d&1&96&Type 1\&2&SiLU\\
 $128^2\times96$&Upsample&2&32&Type 1\&2&SiLU\\
 $256^2\times32$&Conv2d&1&32&Type 1\&2&SiLU\\
 $256^2\times32$&Upsample&2&16&Type 1\&2&SiLU\\
 $512^2\times16$&Conv2d&1&16&Type 1\&2&SiLU\\
 $512^2\times16$&Conv2d&1&3&Type 1\&2&Sigmoid\\
 \hline
 \end{tabular}
 \end{center}
\end{table}
\begin{figure}[tb]
 \includegraphics[width=0.98\linewidth]{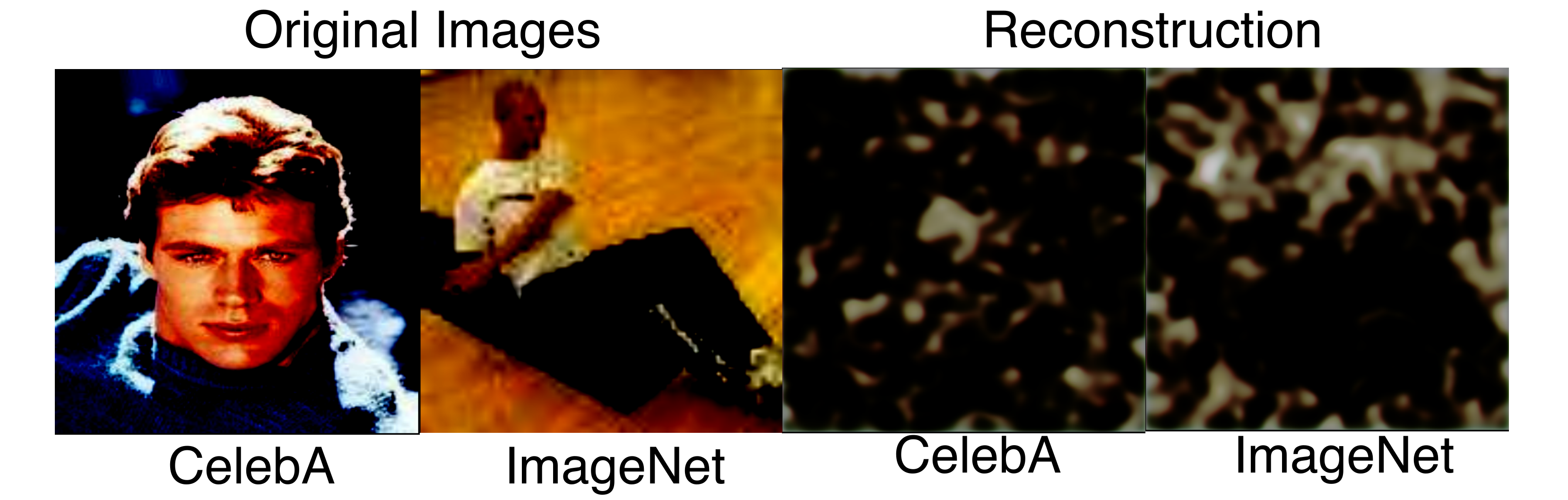}
 \caption{\textbf{Privacy preserving:} Higher distortion means better privacy preservation. Randomly selected and non-cherry-picked examples of reconstructed images from the outputs of the SD Encoder Block 1 using an inverse network.}
 \label{fig:inverseorg}
\end{figure}

For this attack, we examine two aspects: reconstructing the original image and the condition image. Since the server knows the stable diffusion model, it can directly use it to train an inverse network. The key question is how similar private and public datasets are. We choose MS-COCO as the public dataset and CelebA \cite{celeba} and ImageNet \cite{deng2009imagenet} as the private datasets. MS-COCO and ImageNet contain wild images, while CelebA comprises over 200K faces from more than 10K celebrities.

The inverse network is trained on the public dataset and then evaluated on the private dataset. We use the AdamW optimizer with a learning rate of $1\times10^{-5}$, a batch size of 8, and train it for $7.5\times10^{4}$ iterations. The structure of the inverse network is shown in \cref{tab:inverse}. As illustrated in \cref{fig:inverseorg}, this attack is ineffective on both datasets. For CelebA, the PSNR is 5.87 and SSIM is 0.73, while for ImageNet, the PSNR is 6.56 and SSIM is 0.71. The reconstruction results are hardly recognizable as the original private images seen by human eyes.

Secondly, for the reconstruction of the condition image, if we employ our proposed structure, the server can directly train the inverse network. However, in the original structure, the server first utilizes its model weights to train an estimated client model on the public dataset and then trains the inverse network. Unfortunately, this attack is effective for both structures when considering condition images. We will present the results and defense mechanisms in the following sections.

\subsection{Summary}
We summarize potential split learning attacks in \cref{tab:attacks} and their effectiveness. The remaining effective method is inverse network-based attacks for reconstructing condition images. Another valid threat is the leakage of text prompts.

\begin{table*}[tb]
 \caption{Summary of existing attacks in split learning, assessing validity and effectiveness in the diffusion model scenario, using $\checkmark$ for successful data reconstruction and $\times$ otherwise; -- denotes N/A.}
 \label{tab:attacks}
 \begin{center}
 \begin{tabular}{|c|c|cccccc|}
 \hline
 & & \multicolumn{3}{c|}{Original structure} & \multicolumn{3}{c|}{Our structure} \\ \hline
 & & \multicolumn{1}{c|}{Valid?} & \multicolumn{1}{c|}{Raw image} & \multicolumn{1}{c|}{Condition image} & \multicolumn{1}{c|}{Valid?} & \multicolumn{1}{c|}{Raw image} & Condition image \\ \hline
 \multirow{3}{*}{Gradient descent} & White-box & \multicolumn{1}{c|}{$\checkmark$} & \multicolumn{1}{c|}{$\times$} & \multicolumn{1}{c|}{--} & \multicolumn{1}{c|}{$\checkmark$} & \multicolumn{1}{c|}{$\times$} & $\times$ \\ \cline{2-8} 
 & Query-based & \multicolumn{1}{c|}{$\times$} & \multicolumn{1}{c|}{--} & \multicolumn{1}{c|}{--} & \multicolumn{1}{c|}{$\times$} & \multicolumn{1}{c|}{--} & -- \\ \cline{2-8} 
 & Black-box & \multicolumn{1}{c|}{$\checkmark$} & \multicolumn{1}{c|}{--} & \multicolumn{1}{c|}{$\times$} & \multicolumn{1}{c|}{$\times$} & \multicolumn{1}{c|}{--} & -- \\ \hline
 \multirow{2}{*}{Inverse network} & White-box & \multicolumn{1}{c|}{$\checkmark$} & \multicolumn{1}{c|}{$\times$} & \multicolumn{1}{c|}{--} & \multicolumn{1}{c|}{$\checkmark$} & \multicolumn{1}{c|}{$\times$} & $\checkmark$ \\ \cline{2-8} 
 & Black-box & \multicolumn{1}{c|}{$\checkmark$} & \multicolumn{1}{c|}{--} & \multicolumn{1}{c|}{$\checkmark$} & \multicolumn{1}{c|}{$\times$} & \multicolumn{1}{c|}{--} & -- \\ \hline
 Label leakage & -- & \multicolumn{6}{c|}{Invalid: only applicable to binary image classification.} \\ \hline
 Inference attack & -- & \multicolumn{6}{c|}{Invalid: detectable as the model cannot generate the correct results.} \\ \hline
 Text prompt leakage & -- & \multicolumn{6}{c|}{The assumption is valid.} \\ \hline
 \end{tabular}
 \end{center}
\end{table*}

\section{Privacy-Preserving Training of ControlNet}
\subsection{Local Differential Private Timestep Sampling}
We have empirically shown that ControlNet itself is quite effective in defending against several attacks in split learning. If we look at its structure carefully, we will find that the forward process in \cref{fig:flowdiffusion} has already contained the process of adding noise over the latent representation. We will next show that such a mechanism is $(\epsilon,\Delta)-$LDP using the \cref{def:eLDP}. Based on this property, we propose a new sampling scheme over timesteps during the diffusion process, preserving privacy.

With a given latent representation $z_0$, we will generate the noisy latent representation according to the timestep $t$, scheduling parameter $\beta_t$ and a randomly generated noise $\hat{n_t}\sim\mathcal{N}(0,1)$:
\begin{equation}
 \label{eqa:forward}
 \begin{aligned}
 &z_t=\sqrt{1-\beta_t}z_0+\sqrt{\beta_t}\hat{n_t}\\
 &\frac{z_t}{\sqrt{1-\beta_t}}=z_0+\sqrt{\frac{\beta_t}{1-\beta_t}}\hat{n_t}
 \end{aligned}
\end{equation}

According to \cref{fig:cutcontrol}, $z_t$ is the input to the first encoder block of stable diffusion model. Because $\beta_t$ is usually a small number, we approximate \cref{eqa:forward} as, 
\begin{equation}
 z_t\approx z_0+\sqrt{\frac{\beta_t}{1-\beta_t}}\hat{n_t}
\end{equation}
We can view this equation as adding a noise following distribution $\mathcal{N}(0,\frac{\beta_t}{1-\beta_t})$ over $z_0$ to get $z_t$. We then substitute the variance in \cref{def:eLDP}. For convenience, we notate $H$ as hyper-parameter $2\ln\frac{1.25}{\Delta}\alpha^2$.
\begin{equation}
 \begin{aligned}
 &\frac{\beta_t}{1-\beta_t}=H\cdot\frac{1}{\epsilon^2}\\
 &\epsilon=\sqrt{H\cdot\frac{1-\beta_t}{\beta_t}}
 \end{aligned}
\end{equation}
In the diffusion model, we employ the linear scheduling as the default method which is $\beta_t=k\cdot t+\beta_0$, where $k$, $\beta_0$ are scheduling parameters. Hence, we can derive the following relationship between privacy budget $\epsilon$ and $t$, $k$, and $\beta_0$:
\begin{equation}
 \label{eqa:diffusiondp}
 \epsilon(t,k,\beta_0)=\sqrt{H\cdot\left(\frac{1}{kt+\beta_0}-1\right)}
\end{equation}

From this equation, we can see that the privacy budget is related to the timestep. However, to protect different types of images, we need to set different levels of privacy budgets. For example, during the inversion attack for the original image, even if the $t=0$, privacy can still be protected. However, for condition images which are simpler than detailed natural images, the budget at $t=0$ is not enough to protect privacy. In such a case, we need to set bigger privacy budgets. Fortunately, based on \cref{eqa:diffusiondp}, we can set proper privacy budget by setting different $t,k,\beta_0$ in fine-tuning ControlNet.
\begin{theorem}
 \label{theorem:ldp}
 ($(\epsilon,\Delta)$-$\rm{LDP}$ timestep sampling mechanism in diffusion model) With a given privacy budget $\epsilon$, we can have a sampling process in diffusion model which is $(\epsilon,\Delta)$-$\rm{LDP}$. The value of $\epsilon$ is set by a timestep ranging in $[t_s,t_{\max}]$ and scheduling parameters $k$ and $\beta_0$, according to \cref{eqa:diffusiondp}.
\end{theorem}
\subsection{Noise-Confounding Activation Function}
However, as we can see from the \cref{eqa:forward}, if we directly send the encoded condition mixed with the noisy latent representation to the server, as the server knows the timestep $t$ and the label $\hat{n}$, it can directly subtract the added noise from \cref{eqa:forward}. As a result, based on observation of the success defense by the diffusion model part, we find that leveraging some functions, especially non-linear, after the noisy latent representation and before sending to the server will confuse the attacker from stealing privacy. Hence, we propose to add a noise-confounding activation layer before sending features to the server. To design an activation function keeping privacy-preserving property while maintaining the image generation performance, we pass the sum of the encoded condition and the noisy latent representation through such a function:
\begin{equation}
 \label{eqa:activation}
 y=\lvert x \rvert \cdot\left(\frac{2}{1+e^{-x}}\right)+\delta
\end{equation}

\begin{figure}[tb]
 \begin{minipage}{0.38\linewidth}
 \centerline{\includegraphics[width=\linewidth]{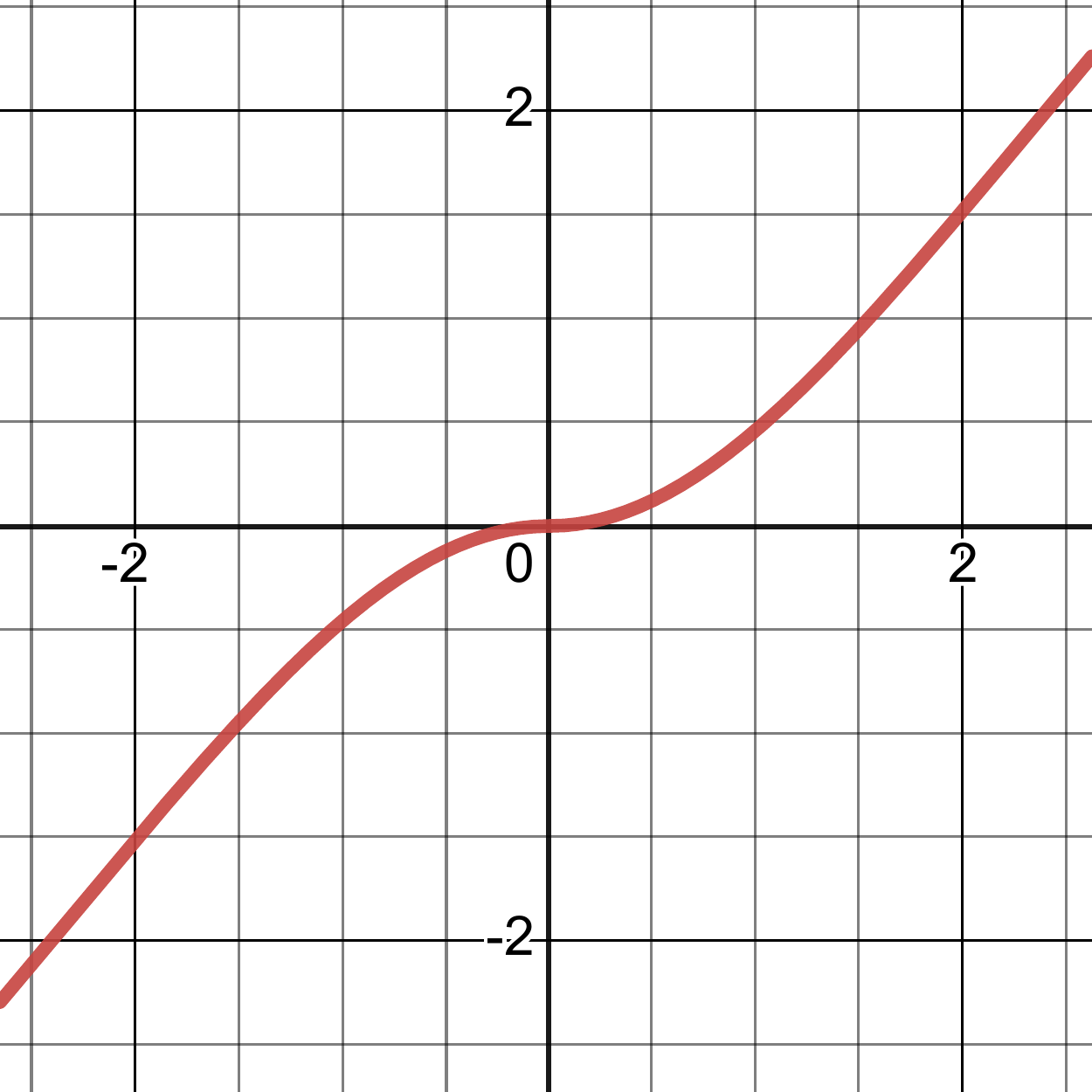}}
 \caption{The graph of \cref{eqa:activation} when $\delta=0$.}
 \label{fig:act}
 \end{minipage}
 \hfill
 \begin{minipage}{0.55\linewidth}
 \includegraphics[width=\linewidth]{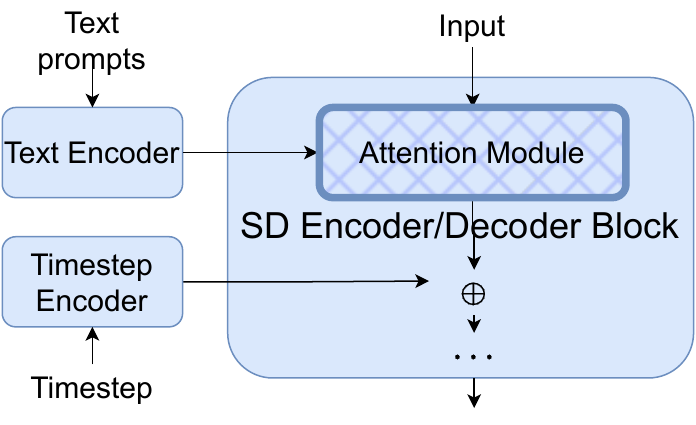}
 \caption{The structure of the encoder and decoder block in original stable diffusion model and ControlNet.}
 \label{fig:block}
 \end{minipage}
\end{figure}

The $\delta$ is a randomized noise following distribution $\mathcal{N}\sim(0,1)$. The noise $\delta$ is randomized at the beginning of the training and fixed during the training. The server or the attacker has no access to $\delta$. The function graph of \cref{eqa:activation} is shown in \cref{fig:act}. The functionality of this function is to prevent the attacker from inferring the sum of the latent representation and encoder condition. In order to maintain image generation performance, we adopt a symmetric design with an SiLU-like shape. The SiLU function~\cite{hendrycks2016gaussian} is a widely used activation function, which can help improve the performance of neural networks. As $\delta$ is fixed during the training, the quality of the summation will not be degraded.

Another solution is to put the SD Encoder Block 1 of the control network on the clients. However, such moving will increase the computation overhead on the clients and break the structure of gradient back-sending free. On the other hand, adding this activation function will help preserve privacy without adding any overhead. Besides, moving the place of SD encoder block is specific for ControlNet while adding activation function can also apply to other conditional diffusion model such as T2I-Adapter~\cite{t2i}.

\subsection{Prompt-Hiding Training}
Apart from keeping the privacy of images, text prompts can also contain private information. In the original ControlNet, for each encoder block and decoder block in the diffusion model and control network, text prompts will be input into the blocks and attention modules will be applied in these blocks to let the generate images learn the text prompts. As a result, the clients have to upload the text prompts or the text features to the server. The server can get the raw text information and privacy will be leaked. Uploading the output features of text encoder may be able to let the server not get texts, but the server can still use the same method proposed by Carlini et al.~\cite{nicolas2023extracting} to extract training dataset using text features as inputs.

As a result, to hide prompts from the server, we propose to not send the text prompts to the server directly. During the training, only the SD Encoder Block 1 in~\cref{fig:cutcontrol} on the clients, will take in text prompts as the input. The text prompts will not be uploaded to the server. Therefore, other encoder and decoder blocks in ControlNet, situated on the server, won't utilize text prompts as inputs. Removing text prompts will not affect the performance of condition and image encoders as they are irrelevant to prompts. However, the input distribution of server-side encoders and decoders has changed. To maintain high-quality image generation, we introduce the following prompts-hiding training methods.

As the diffusion model is frozen and able to always keep the generation performance of a well-trained diffusion model while the control network needs further training, we use different policies for the encoder and decoder blocks in the control network and diffusion model. For blocks in control networks, no text features will be input and the attention modules will be replaced by self-attention modules for the condition features. Since we still need to further fine-tune the control network, the distribution drift caused by removal of text input will be mitigated during training.

To maintain the high image generation performance of the frozen diffusion model, we will keep the text attention modules but input a zero text feature. The zero text feature has the same feature dimension (by default 768) as the original text feature but with a length of one and always has the weights of zero. We need to keep the distribution of text input the same for these blocks as they will not be further trained. Otherwise distribution drift will affect image generation performance.
\section{Evaluation}

\begin{table*}[tb]
	\caption{Comparison of image generation and privacy preservation among different methods, where arrow directions indicate superior image quality and increased difficulty in recognizing reconstructed data.}
	\label{tab:mainresult}
	\small
	\begin{center}
		\begin{threeparttable}
			\begin{tabular}{cccccccccccc}
				\hline
				\hline

				\multicolumn{12}{c}{Defendable by our structure without privacy-preserving methods}                                                                                                                                                                                                                                                                                                           \\
				\hline
				\multicolumn{1}{c|}{Condition} & \multicolumn{6}{c|}{Scribble}                                                                                                                                  & \multicolumn{5}{c}{Segmentation}                                                                                                                \\
				\hline
				\multicolumn{1}{c|}{ \multirow{3}{*}{Methods}}          & \multicolumn{2}{c|}{Performance}                                  & \multicolumn{4}{c|}{Privacy}                                                                   & \multicolumn{5}{c}{Privacy}                                                                                                                 \\
				\multicolumn{1}{c|}{}          & \multirow{2}{*}{FID $\downarrow$} & \multicolumn{1}{c|}{\multirow{2}{*}{CLIP $\uparrow$}} & \multicolumn{2}{c}{CelebA}                  & \multicolumn{2}{c|}{Imagenet}                    & \multicolumn{5}{c}{Imagenet}                                                                                                                \\
				\multicolumn{1}{c|}{}          &                      & \multicolumn{1}{c|}{}                      & PSNR $\downarrow$                 & SSIM $\downarrow$                & PSNR $\downarrow$                & \multicolumn{1}{c|}{SSIM $\downarrow$} & \multicolumn{3}{c}{PSNR $\downarrow$}                                                                 & \multicolumn{2}{c}{SSIM $\downarrow$}                         \\
				\hline
				\multicolumn{1}{c|}{Centralized}  &19.53  & \multicolumn{1}{c|}{26.04}                      & --            & --         & --   & \multicolumn{1}{c|}{--}   & \multicolumn{3}{c}{--}                                                      & \multicolumn{2}{c}{--}                    \\
				\multicolumn{1}{c|}{SL}      &  19.46                    & \multicolumn{1}{c|}{26.87}                      &14.41&0.37&8.17& \multicolumn{1}{c|}{0.35}     & \multicolumn{3}{c}{11.53}                                                        & \multicolumn{2}{c}{0.50}                      \\
				\multicolumn{1}{c|}{Ours}      &  13.45                    & \multicolumn{1}{c|}{26.85}                      &13.15&0.37&7.34& \multicolumn{1}{c|}{0.47}     & \multicolumn{3}{c}{9.95}                                                        & \multicolumn{2}{c}{0.47}                      \\
				
				\hline
				\hline
				\multicolumn{12}{c}{Not defendable by our structure without privacy-preserving methods}                                                                                                                                                                                                                                                 \\
				\hline
				\multicolumn{1}{c|}{Condition} & \multicolumn{6}{c|}{Canny}                                                                                                        & \multicolumn{4}{c|}{Segmentation}                                                                        & \multirow{4}{*}{ \begin{tabular}[c]{@{}c@{}}Attack\\works?\end{tabular}} \\ \cline{1-11}
				\multicolumn{1}{c|}{ \multirow{3}{*}{Methods}}          & \multicolumn{2}{c|}{Performance}                                  & \multicolumn{4}{c|}{Privacy}                                  & \multicolumn{2}{c|}{Performance}                                  & \multicolumn{2}{c|}{Privacy}     &                                \\
				\multicolumn{1}{c|}{}          & \multirow{2}{*}{FID $\downarrow$} & \multicolumn{1}{c|}{\multirow{2}{*}{CLIP $\uparrow$}} & \multicolumn{2}{c}{CelebA} & \multicolumn{2}{c|}{Imagenet}    & \multirow{2}{*}{FID $\downarrow$} & \multicolumn{1}{c|}{\multirow{2}{*}{CLIP $\uparrow$}} & \multicolumn{2}{c|}{CelebA}      &                                \\
				\multicolumn{1}{c|}{}          &                      & \multicolumn{1}{c|}{}                      & PSNR $\downarrow$        & SSIM $\downarrow$        & PSNR $\downarrow$ & \multicolumn{1}{c|}{SSIM $\downarrow$} &                      & \multicolumn{1}{c|}{}                      & PSNR $\downarrow$ & \multicolumn{1}{c|}{SSIM $\downarrow$} &                                \\                      
				
				\hline
				\multicolumn{1}{c|}{Centralized}&11.60&\multicolumn{1}{c|}{26.42}&--&--&--&\multicolumn{1}{c|}{--}&15.23&\multicolumn{1}{c|}{26.82}&--&\multicolumn{1}{c|}{--}&$\checkmark$\\
				\multicolumn{1}{c|}{SL}&11.46&\multicolumn{1}{c|}{26.61}&18.54&0.89&23.10&\multicolumn{1}{c|}{0.94}&17.74&\multicolumn{1}{c|}{27.76}&11.80&\multicolumn{1}{c|}{0.45}&$\checkmark$\\
				\hline
				\multicolumn{1}{c|}{Ours}      &     18.59                 & \multicolumn{1}{c|}{26.21}                      & \multirow{2}{*}{18.86} & \multirow{2}{*}{0.73} & \multirow{2}{*}{22.84} & \multicolumn{1}{c|}{\multirow{2}{*}{0.86}} &14.35& \multicolumn{1}{c|}{26.92}                      & \multirow{2}{*}{12.18} & \multicolumn{1}{c|}{\multirow{2}{*}{0.49}} & \multirow{2}{*}{$\checkmark$}  \\
				\multicolumn{1}{c|}{Ours+t}    & 16.80                & \multicolumn{1}{c|}{26.20}                 &                        &                       &                        & \multicolumn{1}{c|}{}                      &  15.68                    & \multicolumn{1}{c|}{26.70}                      &                        & \multicolumn{1}{c|}{}                      &                                \\
				\multicolumn{1}{c|}{Ours+c}    & 14.52                & \multicolumn{1}{c|}{26.80}                 & \multirow{2}{*}{17.45} & \multirow{2}{*}{0.51} & \multirow{2}{*}{21.74} & \multicolumn{1}{c|}{\multirow{2}{*}{0.70}} &   15.05                   & \multicolumn{1}{c|}{26.85}                      & \multirow{2}{*}{1.68}      & \multicolumn{1}{c|}{\multirow{2}{*}{0.46}}     & \multirow{2}{*}{$\times$}      \\
				\multicolumn{1}{c|}{Ours++}    & 16.80                & \multicolumn{1}{c|}{26.50}                 &                        &                       &                        & \multicolumn{1}{c|}{}                      & 16.32                     & \multicolumn{1}{c|}{26.39}                      &                        & \multicolumn{1}{c|}{}                      &                               
				\\       

				\hline
				\multicolumn{1}{c|}{LDP rr}&18.11&\multicolumn{1}{c|}{27.22}&18.86&0.82&23.77&\multicolumn{1}{c|}{0.97}&17.49&\multicolumn{1}{c|}{27.23}&14.92&\multicolumn{1}{c|}{0.72}&$\checkmark$\\
				\multicolumn{1}{c|}{LDP 0.1}&18.00&\multicolumn{1}{c|}{27.15}&16.84&0.03&19.70&\multicolumn{1}{c|}{0.04}&17.96&\multicolumn{1}{c|}{27.15}&7.56&\multicolumn{1}{c|}{0.33}&$\times$\\
				\multicolumn{1}{c|}{LDP 0.3}&17.28&\multicolumn{1}{c|}{27.12}&18.65&0.79&23.33&\multicolumn{1}{c|}{0.88}&17.21&\multicolumn{1}{c|}{27.13}&8.41&\multicolumn{1}{c|}{0.36}&$\checkmark$/$\times$\\
				\multicolumn{1}{c|}{LDP 0.5}&12.27&\multicolumn{1}{c|}{26.53}&19.81&0.90&24.31&\multicolumn{1}{c|}{0.95}&17.46&\multicolumn{1}{c|}{27.21}&11.21&\multicolumn{1}{c|}{0.51}&$\checkmark$\\
				\multicolumn{1}{c|}{Add 1}&11.77&\multicolumn{1}{c|}{26.60}&25.69&0.98&31.02&\multicolumn{1}{c|}{0.995}&17.51&\multicolumn{1}{c|}{27.30}&22.96&\multicolumn{1}{c|}{0.88}&$\checkmark$\\
				\multicolumn{1}{c|}{Add 50}&19.69&\multicolumn{1}{c|}{26.84}&25.53&0.99&30.60&\multicolumn{1}{c|}{0.99}&17.60&\multicolumn{1}{c|}{27.29}&23.05&\multicolumn{1}{c|}{0.90}&$\checkmark$\\
				\multicolumn{1}{c|}{Mixup}&401.62&\multicolumn{1}{c|}{13.54}&17.96&0.14&22.84&\multicolumn{1}{c|}{0.19}&384.24&\multicolumn{1}{c|}{13.99}&13.45&\multicolumn{1}{c|}{0.73}&$\times$/$\checkmark$\\
				\multicolumn{1}{c|}{PS}&17.39&\multicolumn{1}{c|}{27.16}&21.25&0.95&25.85&\multicolumn{1}{c|}{0.98}&17.62&\multicolumn{1}{c|}{27.22}&22.64&\multicolumn{1}{c|}{0.92}&$\checkmark$\\
				\multicolumn{1}{c|}{FedAvg}&19.10&\multicolumn{1}{c|}{26.92}&--&--&--&\multicolumn{1}{c|}{--}&17.51&\multicolumn{1}{c|}{27.26}&--&\multicolumn{1}{c|}{--}&$\times$\\
				\hline 
			\end{tabular}
			\begin{tablenotes}
				\item  {\small For privacy: $\checkmark$ and $\times$ whether the attack is able to reconstruct condition image. -- means not applicable.}
			\end{tablenotes}
		\end{threeparttable}
	\end{center}
\end{table*}

\subsection{Experimental Settings}
We first introduce details about experimental settings and then evaluate the effectiveness of our methods and state-of-the-art mechanisms defending against successful inversion attacks. We conduct all our experiments on \textsc{Plato}\cite{plato}, an open-source research framework for deploying decentralized training on multiple devices. \textsc{Plato} can support large-scale decentralized training and help deploy the server and the clients on separate devices conveniently. We use the same setting as previous where we have 50 clients in total and each has 1000 training samples. The number of clients will effect efficiency and scalability but will not effect image generation performance or privacy-preserving ability. Since the main focus of this paper is on the latter two aspects, we do not particularly study other settings.

For the pre-trained models, we used stable diffusion V-1.5 and ControlNet V-1.1. The autoencoder is from pre-trained CLIP model with ViT-Large-Patch14~\cite{clip}. The resolution of the input and generate images is $512\times512$. We used the MS-COCO~\cite{coco} as the training dataset for fine-tuning diffusion models to generate high quality images with given conditions. The MS-COCO dataset contains over 120K wild images with proper prompts. It is a common dataset used in fine-tuning large diffusion models and text-to-image generation tasks. The model is fine-tuned over MS-COCO for 25000 iterations with a batch size of 4. The rest training settings is the same as default implementation of ControlNet~\cite{controlnet} where we use AdamW optimizer with learning rate of $1\times10^{-5}$. The noise coefficient $\lambda_t$ is 0.

\textbf{Evaluation Metrics.} For comparing the performance, we need to verify that the privacy-preserving method will not harm image generation performance and that an adversary will not be able to reconstruct private images. For the first objective, we use Fréchet Inception Distance~\cite{fid} (FID) to evaluate quality of generated images and the CLIP score~\cite{clip} to evaluate whether the prompts and generated images are matched (in range of $[0,100]$). We use the MS-COCO validation set with over 5000 images to evaluate the quality of generated images. Lower FID indicates better quality of generated images. A higher CLIP score indicates that the text prompts and the generated images match to each other better. 

For the second prospective, we use PSNR and SSIM as mentioned in previous sections. The images are generated with the same random seed. The settings for evaluating privacy against reconstructing private data is the same as in \cref{sec:privacy}. Lower PSNR and SSIM indicate the reconstructed images are less similar to the private data, meaning better privacy-preserving effectiveness. 

For inverse-network based attacks, if we use the original split learning structure, the attack is under a black-box setting. If we use our designed split learning structure, as the weights of models on clients only involve weights of pre-trained CLIP models downloaded from~\cite{clipdownload}, the attack is under a white-box setting. Within the realm of conditional image generation, various tasks involve different conditions. We assess three types of conditions: canny lines, scribbles, and segmentation maps. These conditions represent a range from detailed to coarse-grained, with the lines drawn in the condition images varying accordingly. 

During the training process of ControlNet, the timestep is sampled among the range of $[t_s,t_{\max}]$. With the default $k$ and $\beta_0$, we say $\epsilon_s=\epsilon(t_s,k,\beta_0)$. So, according to \cref{eqa:diffusiondp}, during the training process, the privacy budget is equal to or larger than $\epsilon_s$. Because we need to ensure that the privacy of every image is preserved, when we evaluate the effectiveness of privacy-preserving methods, in terms of both numerical data and visualization, we consider the worst case of least noise added and sample the timestep as $t_s$ and send the intermediate features to the server. 

\subsection{Implementation of Our Methods and Baselines}
For our and other privacy-preserving methods, we implement them with our designed gradient sending-back free structure. For $(\epsilon,\Delta)$-LDP mechanism, $\Delta=1\times10^{-4}$ and we calculate that $\alpha\approx0.16$. The training latency remains the same after adding our privacy-preserving methods. 

\subsubsection{Implementation of our methods} For our privacy preserving methods, we set the $t_{\max}$, $k$, and $\beta_0$ as default in ControlNet~\cite{controlnet} which are $1000$, $1.115\times10^{-5}$ and $8.85\times10^{-4}$ respectively. If $t_s$ is too big, the sampling range will be too small to get enough samples. If $t_s$ is too small, no privacy protection will be guaranteed. Hence, we set the $t_s$ around middle point which is 536, which results in $\epsilon_s\approx 8$. We implement our methods in three ways: only protecting conditions, only hiding prompts and both. We denote our structure without any privacy-preserving methods implemented (\cref{sec:ours} ) as \textbf{Ours}. We denote our three ways of implementation as \textbf{Ours+c}, \textbf{Ours+t}, and \textbf{Ours++}, respectively.

\subsubsection{Implementation of baselines}
We compare our methods with several state-of-the-art privacy-preserving methods which can be used for split learning with ControlNet. The rationale for showing the results with the following chosen parameters is that we would like to show the cases that the baselines neither protect privacy nor generate images of high quality. If they want to generate a good image, they should choose a smaller disruption magnitude (e.g. smaller privacy budget). This will make privacy-preserving performance weaker. On the other hand, if they want to provide stronger privacy protection, they need to use stronger disruption, which will further degrade image generation performance. These can show that they cannot find a proper solution for preserving privacy and image generation performance at the same time.

\textbf{LDP rr} means a mechanism called randomized response which is local differential private~\cite{ldprr}. We implement randomized response over intermediate results following steps in literature where we set privacy budget as $2$.

\textbf{LDP number} means the mechanism in \cref{def:eLDP}. The number means the privacy budgets where we have three values: $0.1$, $0.3$, and $0.5$. 

\textbf{Add number} means the mechanism of adding Gaussian noise on the raw data according to the distribution $\mathcal{N}\sim(0,\sigma^2)$. We have two numbers: $\sigma^2=1$ and $\sigma^2=50$. 

\textbf{Mixup} is the method of mixing up data proposed in DataMix ~\cite{datamix} and CutMix~\cite{oh2022differentially}. We mix four images together which is the same as the batch size.

\textbf{PS} is the method called patch shuffling \cite{patchshuffling,xu2023shuffled}. The patch size is set to 4, same as the batch size. 

We also compare the results of image generation with several other baselines.

\textbf{Centralized} means images generated by the well-trained ControlNet V 1.1 from~\cite{trainedcontrolnet}. We directly use the downloaded models to generate images. This is a production-level baseline.

\textbf{SL} is the deployment of ControlNet with split learning without any privacy-preserving methods applied (\cref{sec:sl}). We fine-tune ControlNet following steps of split learning.
\subsection{Comparison Results}
We present qualitative results in \cref{tab:mainresult}. The conclusion is that from numerical data and visualization, we can see that \textbf{our method is the only method that can protect privacy without loss of image generation quality.} The methods that can generate images correctly cannot preserve privacy well. The methods can preserve privacy well are not able to generate good images. Though more advanced methods such as Mixup and PS can provide good privacy protection in some cases, they fail to correctly generate images of good quality conforming to the conditions.

One of our new insights is that all previous privacy-preserving methods try to propose a general method for split learning, overlooking the variance between different use cases. We can easily extend methods like DataMix from image classification to different tasks. However, they cannot achieve satisfactory performance when we really verify them on the task of image generation. \textbf{Our privacy-preserving method is tailored for diffusion models}, considering the specialty of the overall model structure of diffusion model to how prompts are processed. Let's take a look at the detailed analysis.

\subsubsection{Maintenance of image generation performance }

An interesting result is that our designed split learning structure not only improves the efficiency but also improves the quality of generated images, reflected by FID. For example, on scribble conditions, we can improve FID from 19.53 to 13.45. This is possible as the pre-trained CLIP model is well-trained on large datasets. While for other methods, such as LDP Gaussian noise adding, though they can provide strong privacy protection with a small privacy budget, they need to sacrifice the image generation quality. Furthermore, our methods preserve data privacy regardless the number of samples on each client. Clients only need to do inferences with our designed structure. The results of inference are irrelevant to the number of samples passed through the models. Another reason is that our methods do not mix several training samples like what Mixup and Patch Shuffling did.

\subsubsection{Privacy-preserving ability}
For canny conditions, attackers find it easier to reconstruct private data, while for scribble conditions, it is much harder. However, as canny contains richer information, including complex lines, protecting such conditions is crucial. \textbf{We analyze privacy concerns for individual conditions and private datasets separately.} In segmentation, attack success depends on the datasets. In cases where split learning with the original and our designed structure can defend against existing attacks, our methods can enhance privacy. For other cases, having risks of data leakage, our methods can protect privacy. For instance, we can reduce PSNR from 11.80 to 1.68.

\begin{figure}[tb]
 \subfigure[$k$, $\epsilon_s=2$]{\includegraphics[width=0.24\linewidth]{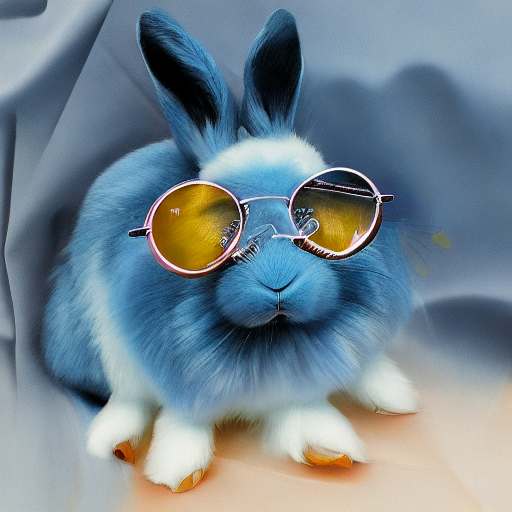}}\hfill
 \subfigure[$k$, $\epsilon_s=0.3$]{\includegraphics[width=0.24\linewidth]{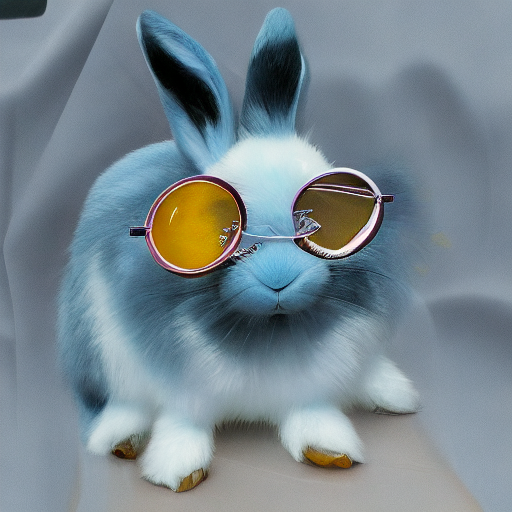}}\hfill
 \subfigure[$\beta_0$, $\epsilon_s=2$]{\includegraphics[width=0.24\linewidth]{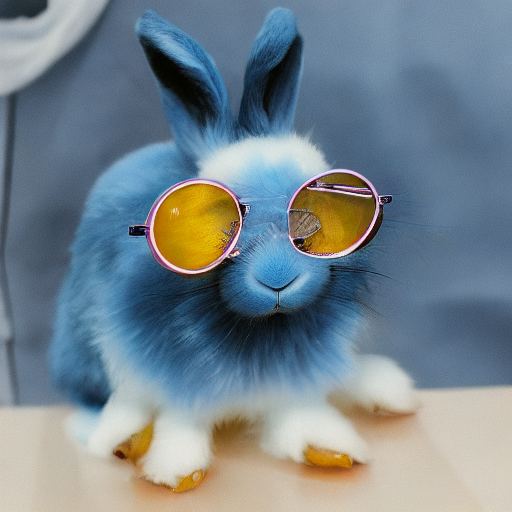}}\hfill
 \subfigure[$\beta_0$, $\epsilon_s=0.3$]{\includegraphics[width=0.24\linewidth]{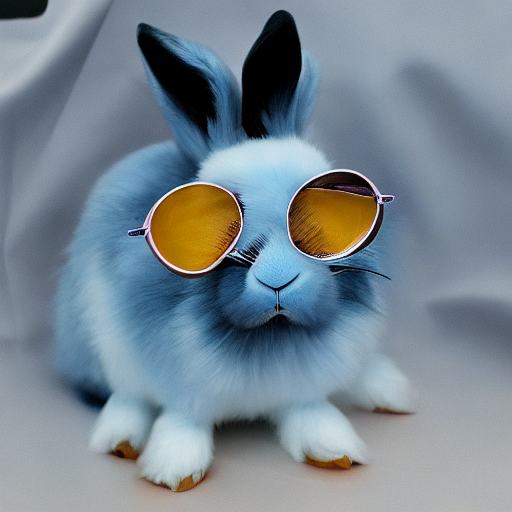}}
 \caption{Randomly selected and non-cherry-picked examples of generated images varying privacy budget with different $k$ and $\beta_0$ in our method (Ours++). $k$ and $\beta_0$ indicate that the privacy budgets change by altering only $k$ or $\beta_0$ respectively compared to the default settings.}
 \label{fig:ablation}
\end{figure}

\subsection{Ablation Study of Privacy Budgets}
In \cref{theorem:ldp}, we can set different privacy budgets with proper $k$ and $\beta_0$. In \cref{fig:ablation}, we change privacy budgets by setting different scheduling parameters $k$ and $\beta_0$ respectively. In the default setting of our method, the privacy budget is $8$. We try the other two cases of setting privacy budgets as $0.3$ and $2$. As shown in \cref{fig:ablation}, our method can still generate images of good quality. However, for the baseline methods, they fail to generate good images and protect privacy when they use the same privacy budgets of $0.3$ and $2$.

\section{Conclusion}

In this paper, we address the challenge of fine-tuning ControlNet models with locally distributed data across multiple users, focusing on feasibility and privacy. We initiate the study with federated learning and find that conventional federated learning is not suitable due to high GPU memory requirements during training, the unavailability of stable diffusion models, and empirically proven performance degradation. So we turn to split learning to solve such a problem where we first improve the structure so that the server does not need to send gradients back to the clients, greatly improving efficiency. Through in-depth study of existing attacks in split learning. We discover that the effectiveness of most existing methods is weakened. For the remaining threats, we propose differential private timestep sampling, a noise-confounding activation function, and prompts-hiding training, based on the built-in mechanisms in diffusion models with tunable privacy budgets. We show convincing results from a wide array of experiments that our method can provide stronger privacy protection without loss of image generation performance and train the models faster than its state-of-the-art alternatives in the literature.

\newpage
\bibliographystyle{IEEEtran}
\bibliography{main}

\appendix
\appendices

\section{Discussion}
In this paper, we resolve the question of how we can train ControlNet and diffusion models while keeping users' data privacy. Besides the aspect of preserving privacy, there are other issues worth studying in production level split learning with ControlNet and stable diffusion. Another challenging question is how we can keep users' data privacy during the inference stage after deploying trained ControlNet and diffusion models. The inference process is different from the training. A trivial solution is to run the inference completely on the edge device, which needs about 7.5GB of memory. The memory requirement is much less than that of training, which is feasible. However, maybe not all clients have enough memory. It is a challenge that how we can still keep user data privacy if we deploy a ControlNet across the clients and the server. From related work, we can see large efforts are being put into privacy-preserving inference in split learning. It is worth studying whether these methods are helpful during the inference stage.

In this paper, the target is towards privacy-preserving split learning with ControlNet and diffusion model. In a broader research topic, one question is how we can safely do split learning. In such a case, we may not assume every client is honest, which means some clients are malicious and not sending the correct intermediate features. To harm the interests of other clients, some clients may do backdoor attacks or adversarial attacks, diminishing the utility of the fine-tuned ControlNet and diffusion model. 

In our experiments, we deploy split learning with 50 clients. We can increase the number of clients if we want, but since $T_s$ is much larger than $T_c$, the whole training time is the same. Therefore, we do not increase the number. On the production level, it is possible that there are more than 50 clients. With our methods, we can still train ControlNet with split learning over them while preserving data privacy. A minor issue is that since the clients only need to do inference, they may send intermediate features of large amounts continuously and simultaneously. It is worth studying how the server deals with a large scale of requests simultaneously. We can expand the client number to hundreds or thousands to evluate the scalability.

\end{document}